\documentclass[10pt,twocolumn,letterpaper]{article}

\usepackage{iccv}
\usepackage{microtype}
\usepackage{graphics}
\usepackage{subcaption}
\usepackage{caption}
\usepackage{booktabs} 

\usepackage{makecell}

\usepackage{times}  
\usepackage{helvet} 
\usepackage{courier}  

\usepackage{epstopdf}
\usepackage{graphicx}
\usepackage{rotating,multirow}
\usepackage{amsmath,amssymb} 
\usepackage{color}
\usepackage{wrapfig,lipsum}
\usepackage{tabularx}

\usepackage{algorithm}
\usepackage{algorithmic}
\usepackage[algo2e]{algorithm2e}
\usepackage{mathtools}
\newcommand{\aka}{\textit{a.k.a. }}

\usepackage[pagebackref=true,breaklinks=true,letterpaper=true,colorlinks,bookmarks=false]{hyperref}

\iccvfinalcopy 

\def\iccvPaperID{2023} 

\ificcvfinal\fi

\begin{document}

\title{Unsupervised Person Re-identification via Multi-Label Prediction and Classification based on Graph-Structural Insight}

\author{Jongmin Yu and Hyeontaek Oh\\
Institute for Information Technology Convergence\\
Korea Advanced Institute of Science and Technology (KAIST)\\
291 Daehak-ro, Yuseong-gu, Daejeon 34141, Republic of Korea\\
{\tt\small \{andrew.yu, hyeontaek\}@kaist.ac.kr}
}

\maketitle
\ificcvfinal\thispagestyle{empty}\fi

\begin{abstract}
This paper addresses unsupervised person re-identification (Re-ID) using multi-label prediction and classification based on graph-structural insight. Our method extracts features from person images and produces a graph that consists of the features and a pairwise similarity of them as nodes and edges, respectively.
Based on the graph, the proposed graph structure based multi-label prediction (GSMLP) method predicts multi-labels by considering the pairwise similarity and the adjacency node distribution of each node. The multi-labels created by GSMLP are applied to the proposed selective multi-label classification (SMLC) loss. SMLC integrates a hard-sample mining scheme and a multi-label classification. The proposed GSMLP and SMLC boost the performance of unsupervised person Re-ID without any pre-labelled dataset. Experimental results justify the superiority of the proposed method in unsupervised person Re-ID by producing state-of-the-art performance. The source code for this paper is publicly available on \url{https://github.com/uknownpioneer/GSMLP-SMLC.git}.
\end{abstract}

\section{Introduction}
\label{sec:1}
Person re-identification (Re-ID) have been achieved great success alongside with the development of various deep learning-based methods \cite{krizhevsky2012imagenet,he2016deep,su2017pose} for extracting discriminative features from pre-labelled person images. However, it is basically an expensive job to create a large-scale and well-annotated dataset. Furthermore, there is a possibility of performance degradation caused by the incorrectly labelled images. According to those issues, interests in unsupervised manners for person Re-ID have been increasing.

\begin{figure}[t]
	\centering
	\includegraphics[width=\columnwidth]{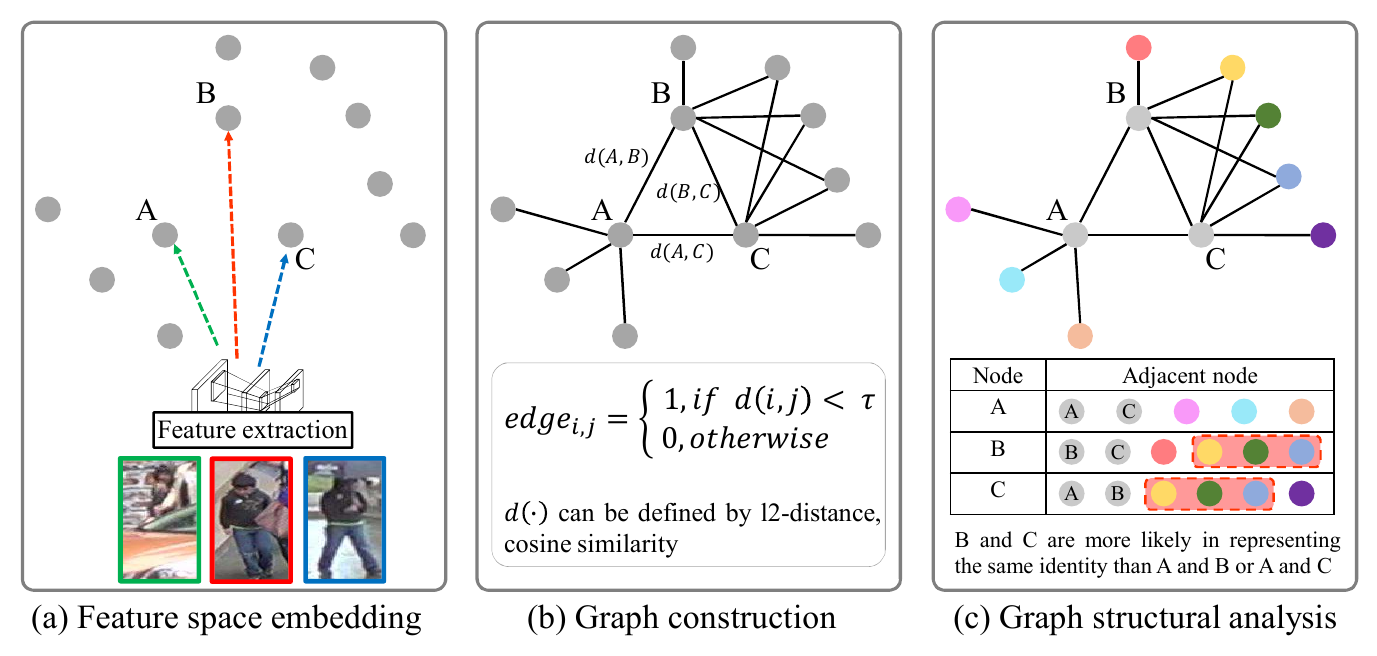}
	\caption{A brief illustration of our graph structure based multi-label prediction (GSMLP). (a) Features are extracted from unlabelled person images and embedded into a latent space. (b) A graph is created by establishing edges that are defined as a distance of node pairs. (c) Our method assigns multi-labels by comparing the adjacent node distribution.}
	\label{fig:brief}
	\vspace{-2ex}
\end{figure}

Learning discriminative features without any pre-labelled images is the primary challenge on unsupervised person Re-ID. A dominant approach is domain adoption (DA) based on transfer learning, which utilises the annotated data from a source domains (\eg other person Re-ID datasets), for parameter initialisation or label transfer. The studies with DA have significantly improved the performance of unsupervised person Re-ID, but they still need labelling tasks for data of the source domain. Additionally, as mentioned in some studies \cite{long2015learning,Yan_2017_CVPR}, finding an optimal source dataset is important in applying DA for unsupervised person Re-ID because the larger domain gap between the source and target domains the more performance degradation. However, it is intractable to measure the domain gap before the end of model training. To avoid this issue, some approaches utilised pseudo labelling techniques based on clustering methods \cite{fan2018unsupervised,Fu_2019_ICCV} or image file-wise indexing \cite{lin2020unsupervised,wang2020unsupervised}. However, the performance of those methods can be affected by the setting of hyper-parameters (\eg the number of cluster and the initial values of centroids).

This paper addresses a multi-label prediction and classification methods for improving the performance of unsupervised person Re-ID without any pre-labelled data or clustering methods. As shown in Figure \ref{fig:brief}, we regard each feature extracted from a single image as a node of a graph. Edges on the graph are defined by a similarity between each feature. Our method predicts pseudo-multi-labels for each image by considering both the pair-wise similarity between each feature and the similarity of neighbourhood node distribution of each feature. The method is trained in terms of minimising a multi-label classification loss. The loss function is formulated not only to minimise the distance of nodes that can be considered as the same identity but also to maximise the distance of nodes that would be regarded as different identities, simultaneously.

The proposed method generates multi-labels about each training data in every training step. Therefore, it is essential to predict faultless multi-labels. To ensure the quality of the multi-label estimation, we propose a Graph Structure-based Multi-Label Prediction (GSMLP) that generates multi-labels based on the graph-structural insight. Straightforwardly, the key hypothesis on GSMLP is that two images can be classified as the same identity if and only if 1) the geometric distance of two nodes (\ie features) is close enough, and 2) adjacent node distributions of the two nodes are structurally similar. To implement this, GSMLP exploits an in-memory based look-up table that stores all extracted features into memory and compares each feature with the features of the look-up table. 

Multi-labels predicted by GSMLP give a chance to apply multi-label classification approaches explicitly. Because our method assumes that each image basically has an independent single class, a conventional multi-label classification, based on a fully connected neural network with a one-hot encoding and softmax function, may not be suitable in terms of model complexity and computational intensity. Moreover, unnecessarily increased model complexity can be sometimes a cause of the curse of dimensionality \cite{40d5d7fd62cb44ba934a8a75d4b2b076}. To overcome this issue, we propose a Selective Multi-Label Classification (SMLC) loss combined with a metric learning methodology and a hard negative sample mining. In our ablation studies, we verify the effectiveness of SMLC compared with the commonly used multi-label classification approach. 

Our method is evaluated on the three publicly available datasets: 1) \textit{Market-1501} \cite{zheng2015scalable}, 2) \textit{DukeMTMC-Re-ID} \cite{ristani2016performance}, and 3) \textit{MSMT17} \cite{wei2018person}. Experimental results demonstrate that the proposed method achieves competitive performance compared with recent state-of-the-art methods. Our method achieves 80.9 and 66.5 of rank-1 accuracies on \textit{Market-1501} and \textit{DukeMTMC-Re-ID}, respectively, and those figures are higher than BUC \cite{lin2019bottom}, DBC \cite{ding12dispersion}, SSL \cite{lin2020unsupervised}, and MMCL \cite{wang2020unsupervised}. Additionally, our method provides comparable performance compared with the recent methods based on DA. Consequently, the proposed method achieves promising performance without any labelled data for person Re-ID, so it shows the possibility that person Re-ID can achieve reasonable performance even if there is no well-labelled dataset. 
 
\begin{figure*}[t]
\vspace{-1ex}
	\centering
	\includegraphics[width=\textwidth]{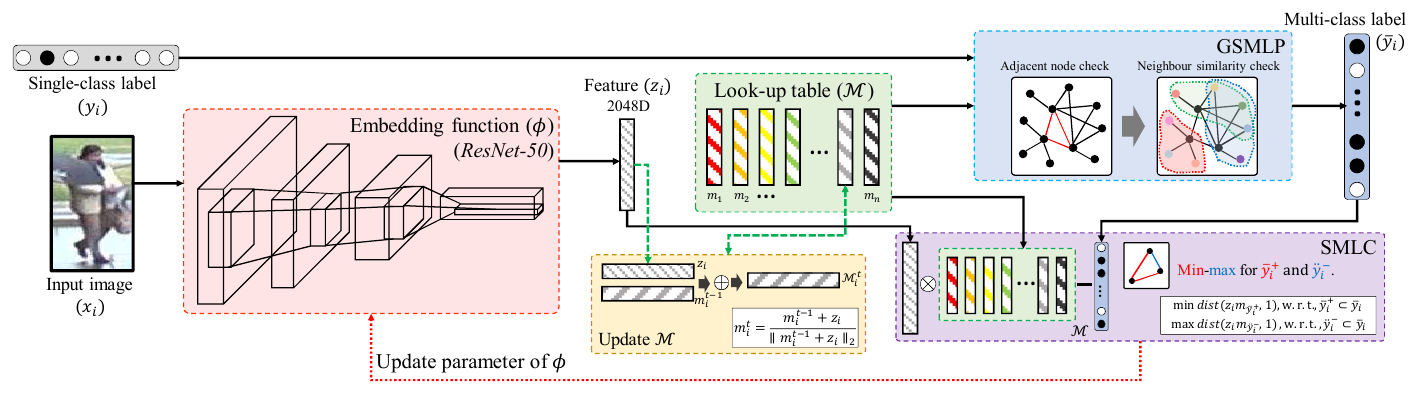}
	\caption{An overview of the proposed method for unsupervised person Re-ID. With a given unlabelled image $x_{i}$, the corresponding pseudo single-class label $y_{i}$ is created. $x_{i}$ is applied to the embedding function $\phi$ to extract the feature $z_{i}$. $z_{i}$ is used to update a look-up table $\mathcal{M}$ or compute a loss. GSMLP predicts a multi-label $\bar{y}_{i}$ using $y_{i}$ and $\mathcal{M}$. SMLC is computed by comparing $\bar{y}_{i}$ and the output $z_{i}\mathcal{M}^{\mathrm{T}}$. Above process is denoted by black coloured solid arrow lines. The dotted arrow lines with red colour show optimisation pipeline for $\phi$. The dashed arrow lines with green colour show the update process of $\mathcal{M}$ using $z_{i}$ and the corresponding element $m_{i}$ in $\mathcal{M}$.}
	\label{fig:proposed_method_training}
	\vspace{-2ex}
\end{figure*}

\section{Related works}
\label{sec:2}
\textbf{Supervised person Re-ID:} A great number of person Re-ID methods have been proposed in a supervision manner, and those methods assumed that a large-scale and well-annotated dataset is given. In the recent decade, with developing of outstanding representation learning method based on deep learning, supervised person Re-ID have been improving rapidly. In particular, convolutional neural networks (CNNs) based methods \cite{zheng2015scalable,zheng2019joint,li2014deepreid,varior2016gated} were proposed to learn discriminative features from person images and achieved remarkable performances. Additionally, recent approaches \cite{he2020guided,chen2020salience,Yang_2019_CVPR} presented prioritised or selective feature learning approaches based on feature saliency map, and those approaches have shown the more improved performance than that of the previous deep learning based methods. 

On the other hand, metric learning has also been frequently used methodology for supervised person Re-ID. The studies \cite{yu2018deep,zhou2020fine,zhou2020online} commonly optimised their models by minimising or maximising a distance of two images depending on whether the images have the same identity or not. These approaches may not need explicit labelling for each person identity, and they just need a label to show that a pair of the image contains the same identity or not. Possibly, metric learning-based approaches can be considered as more flexible solution for supervised person Re-ID.  

However, it is inevitable to assign reliable labels to compile good solutions for person Re-ID in any supervised approaches, and it means that the performance of those methods is variant to the quality of the labels. Accordingly, we propose a fully unsupervised person Re-ID that does not leverage any kind of prepared labels for person images in the training step, in which invariant to the quality of labels.

\textbf{Unsupervised person Re-ID:} Traditional unsupervised methods can be grouped into three categories: 1) leveraging hand-crafted features \cite{liao2015person,zheng2015scalable,gray2008viewpoint}, 2) feature clustering \cite{fan2018unsupervised,Fu_2019_ICCV}, and 3) dictionary learning \cite{BMVC2015_44,li2017person}. Usually, those approaches produced relatively lower performance than that of supervised methods because it is intractable to find the optimal setting for diverse hyper-parameters (\eg the number of clusters), which can not be flexible to numerous situations in the real world (\eg various camera viewpoints and illumination condition change for day and night). 

In recent few years, unsupervised approaches based on DA, which utilise both unlabelled data of a given dataset (\aka the dataset from a target domain) and labelled data of other datasets (\aka the datasets from a source domain), have improved the performance of unsupervised person Re-ID remarkably. Some of them \cite{wang2018transferable,yu2019unsupervised} applied transfer learning with additional attribute annotation for minimising attribute-level domain gaps. However, even if those methods do not need to label the target dataset, those still need pre-acquired annotations from the source domain. Furthermore, when a domain gap (which is hard to predict before conducting actual performance evaluation) between the source and target domains is large, the performance of unsupervised person Re-ID is degraded \cite{long2015learning,wei2018person}. 

To avoid the necessity of any labelled dataset from a source domain, pseudo labelling approaches for the target domain itself have been presented \cite{wang2020unsupervised,lin2020unsupervised}. Lin \etal \cite{lin2020unsupervised} utilised soft labels that do not need a clustering method or hard quantisation for labelling. Wang and Zhang \cite{wang2020unsupervised} proposed a pseudo-multi-label prediction based on the in-memory feature bank. However, these approaches leveraged pairwise information between two images only. In this paper, we propose a multi-label prediction method that considers not only pairwise information between two images but also their neighbourhood information to improve the quality of labelling.

\section{Proposed Method}
\label{sec:3}

\subsection{Methodology} 
\label{sec:3:1}
An overview of the proposed method for unsupervised person Re-ID is illustrated in Figure \ref{fig:proposed_method_training}. The goal of our method is deriving person Re-ID model using unlabelled person images $\mathcal{X}=\{x_{i}\}^{i=1:n}$, where $n$ is the number of the images. 
As similar to the various approaches that generate pseudo labels for unsupervised person Re-ID \cite{lin2020unsupervised,wang2020unsupervised}, the training of our method is also started by initialising single-class labels $y$ to each image $x$ in $\mathcal{X}$ using their indices. Initially, a label $y_{i}\in\mathbb{R}^{n}$ of a training image $x_{i}$ is defined as 
\begin{equation}
y_{i,j} = 
\begin{cases}
1, & j=i, \\
0, & j \neq i,
\end{cases}
\label{eq:initial_label}
\end{equation}
where $y_{i,j}$ indicates $j^{\text{th}}$ element in the label $y_{i}$. The above label definition does not regard the existence of multiple images per one identity, so a different type of label (\ie multi-label) is needed, which can contain correspondence for other images that probably represent the same identity.

In general, with a given query image $q$, person Re-ID models are expected to retrieve an image $\hat{g}$ containing the same person identity from a gallery $\mathcal{G}=\{g_{i}\}^{i=1:m}$, where $m$ is the number of images $\mathcal{G}$. This process is based on the hypothesis that if $q$ and $g$ contain the same identity, the features extracted from two images are more similar than others in $\mathcal{G}$. The above problem can be represented as follows:
\begin{equation}
\hat{g} = \arg\min_{g \in \mathcal{G}} dist(z^{q},z^{g}),
\label{eq:un_Re-ID}
\end{equation}
where $z\in\mathbb{R}^{d}$ is a $d$-dimensional feature vector extracted from $q$ and $g$, and it is normalised by a unit vector $|z|=1$ using $l2$-normalisation. $z$ is obtained by applying images 
($q$ and $g$) to an embedding function $\phi(\cdot)$ ($\phi:x\xrightarrow{}z$). $dist()$ denotes a similarity measuring function.

As Eq. \eqref{eq:un_Re-ID}, a similarity between two images can be used as a likelihood to represent whether two images contain the same identity or not. We adopt this scheme to predict multi-label. To compute pairwise feature similarities across all training images, our method adopts an in-memory look-up table $\mathcal{M}=[m_{i}]^{i=1:n}, \mathcal{M}\in\mathbb{R}^{n\times{}d}$ to store all features, where $m_{i}\in\mathbb{R}^{d}$ is a stored features, and $n$ is number of input images. The look-up table is initialised by using input features $z_{i}$ and continuously updated during the training (see the end of Section \ref{sec:3:1}). 

Using the look-up table $\mathcal{M}$, the similarities between a specific feature ($z_{i}$) and all other features is computed as follows:
\begin{equation}
\begin{aligned}
z_{i}\mathcal{M}^{\mathrm{T}}&=[z_{i,1},z_{i,2},...,z_{i,d}]
\begin{bmatrix}
m_{1,1} & \cdots & m_{n,1} \\
\vdots & \ddots & \vdots \\
m_{1,d} & \cdots & m_{n,d} 
\end{bmatrix}\\&=[z_{i}m^{\mathrm{T}}_{1}, z_{i} m^{\mathrm{T}}_{2}, ..., z_{i} m^{\mathrm{T}}_{n}] = s_{i},  
\label{eq:sim_lookup}  
\end{aligned}
\end{equation}
where $\mathrm{T}$ denotes matrix transpose, $s_{i}\in\mathbb{R}^{n}$ is the feature similarity vector between $z_{i}$ and all other features in $\mathcal{M}$. All features are normalised by unit vectors (\eg $||z|| \approx 1$). Therefore, the inner product between two features is equivalent to computing cosine angular similarity as follows:
\begin{equation}
s_{i,j}=z_{i}\cdot{}m_{j} \equiv \lVert{}z_{i}\rVert{} \lVert{}m_{j}\rVert{}cos\theta_{i,j} \equiv cos\theta_{i,j}, \label{eq:sim_cosine}
\end{equation}
where $\theta_{i,j}$ is an angle between $z_{i}$ and $m_{j}$. If $s_{i,j}$ is closer to 1, then it is more likely that two image $x_{i}$ and $x_{j}$ represent the same identity.

The above process is invariant to both the number of labels and the scalability of a training dataset. However, predicting multi-labels using only a similarity $s$ can generate a lot of false-positive because of the numerous variation on person images. We hence propose a Graph-Structural Multi-Label Prediction (GSMLP) for generating more reliable multi-labels. GSMLP predicts multi-label $\bar{y}_{i}$ by leveraging single-class label $y_{i}$ and $\mathcal{M}$ as follows: 
\begin{equation}
\bar{y}_{i} = \text{GSMLP}(y_{i},\mathcal{M}).
\label{eq:gsmlp}
\end{equation}

$\bar{y}_{i}$ is $n$-dimensional vector:  $\bar{y}_{i}=[y_{i,j}]^{j=1:n}_{y_{i,j}\in[0,1]}$, and if $y_{i,j}$ is 1, than it means $x_{i}$ and $x_{j}$ contain the same identity. GSMLP is a non-parametric method, so multi-label prediction is conducted with $y_{i}$ and $\mathcal{M}$ only. 

The predicted multi-label is applied to the proposed Selective Multi-Label Classification (SMLC) loss. The dominant part of the predicted multi-label $\bar{y}$ is composed by 0, so there is a balancing issue in applying the multi-label to compute the loss. SMLC addresses this issue by employing a selective loss computation that selects and applies particular 0-marked parts based on the similarity score $s$. Computing SMLC is represented by
\begin{equation}
\mathcal{L}_{\text{SMLC}} = \sum_{m=1}^{B}\text{SMLC}(\phi{}(x_{m}) \mathcal{M}^{\mathrm{T}}, \bar{y}_{m}),
\label{eq:SMLC}
\end{equation}
where $\text{SMLC}(\cdot)$ denotes the proposed SMLC module, and $B$ denote the size of a batch for model training. As same as Eq. \eqref{eq:sim_lookup}, $\phi{}(x_{m}) \mathcal{M}^{\mathrm{T}}$ generates a similarity vector $s_{m}$, and it is applied to compute the loss.

The look-up table $\mathcal{M}$ should be updated during model training since extracted features can be varied continuously. Each element of $\mathcal{M}$ is updated as an average of features $z$ and the corresponding features in $\mathcal{M}$ with $l2$-normalisation as follows:
\begin{equation}
m^{t}_{i} \xleftarrow{} \frac{m^{t-1}_{i}+z^{t}_{i}}{\lVert{}m^{t-1}_{i}+z^{t}_{i}\rVert{}_{2}},  \quad \text{s.t. } m^{0}_{i} = z^{0}_{i},
\label{eq:update}
\end{equation}
where $t$ indicates a $t^{\text{th}}$ training step. $l2$-normalisation is adopted to preserve the robustness of GSMLP. Additionally, to reduce the covariate shift \cite{he2016deep,schneider2020improving}, a feature distribution on $\mathcal{M}$ is unstable since $\mathcal{M}$ is not fully updated in each training step, all features on $\mathcal{M}$ is fully re-initialised at a specific point in the training step.

Above series of the process using the proposed GSMLP and SMLC boost the performance of unsupervised person Re-ID by updating the quality of multi-labels and discriminative power of features iteratively and sequentially. The detail explanations about the proposed GSMLP and SMLC are described in the following sections.

\subsection{Graph Structure-based Multi-Label Prediction}
\label{sec:3:2}
With a given single-class label $y$ and the look-up table $\mathcal{M}$, GSMLP focuses on finding multi-labels which can contain information of other images that may have the same identity. To end this, GSMLP plots a graph $G=(V,E)$, where $V$ and $E$ are nodes and edges on the graph, respectively. The node and the edge are defined by the stored features and distances between each features as follows:
\begin{equation}
\begin{aligned}
&V = \{v_{i}|v_{i}\leftarrow{m}_{i} \text{ } \text{w.r.t.} \text{ } 1\leq{}i\leq{}n\}, \\
&E = \{e_{i,j}|e_{i,j}\leftarrow{}dist(v_{i},v_{j}) \text{ } \text{if} \text{ } dist(v_{i},v_{j}) \geq{}\tau\}.
\label{eq:def_graph}
\end{aligned}    
\end{equation}

Similar to \cite{zhang2013review,wang2020unsupervised}, which applied a threshold to find out geometrically near features for each other, in this paper, two nodes are connected if and only if the similarity between two nodes is higher than the predefined threshold $\tau$, so using $E$ can find possible candidates of relevant labels. However, those features may be distributed complicatedly because of numerous visual variations (such as illumination change, the viewpoint of the camera, background noise), and this complication can degrade the quality of multi-labels. To enhance the quality of predicted multi-labels, GSMLP utilises both the connection between two nodes and information about adjacent nodes of the two nodes, simultaneously. This approach is inspired by \cite{jegou2007contextual,zheng2016person} with the following hypothesis: if two images belong to the same identity, their neighbour image sets should be similar. 

Back to the graph definition in Eq. \eqref{eq:def_graph}, $E$ can be estimated by $\mathcal{M} \mathcal{M}^{\mathrm{T}}$ with element-wise filtering using the threshold $\tau$. To represent adjacent node similarity, we define a softened adjacency matrix $\hat{\mathcal{A}}$ as follows:
\begin{equation}
\hat{\mathcal{A}}=\mathcal{M} \mathcal{M}^{\mathrm{T}}=
\begin{bmatrix}
m_{1}\cdot{}m_{1}, \cdots, m_{1}\cdot{}m_{n}\\
\vdots \quad \quad \quad \ddots \quad \quad \quad \vdots \\
m_{n}\cdot{}m_{1}, \cdots, m_{n}\cdot{}m_{n}
\end{bmatrix}=
\begin{bmatrix}
a_{1}\\
\vdots \\
a_{n}
\end{bmatrix}.
\label{eq:estimation}
\end{equation}

Compared with a general adjacent metric (which is a symmetric matrix and represents connectivity on a graph using binary value), $\hat{\mathcal{A}}$ is also symmetric but consists of a real number representing cosine angular similarity between each node. Similar to Eq. \eqref{eq:def_graph}, among the elements in $\hat{\mathcal{A}}$, the elements less than $\tau$ is replaced by 0. From now on, the information for the adjacent images of an image $x_{i}$ can be simply represented by $a_{i}$. 

GSMLP predicts a multi-label using $\hat{\mathcal{A}}$ in two steps. First, with a given single-class label $y_{i}$ as an index, a candidate label list for positive labels is given by 
\begin{equation}
\text{P}^{+}_{i} = \left\{ j  | a_{i,j}\neq{}0, 1 \leq j \leq n,\right\},
\label{eq:first} 
\end{equation}
where $\text{P}^{+}_{i}$ is a set of positive label candidates for the image $x_{i}$ taking the single-class label $y_{i}$. 

Second, as mentioned by \cite{zhang2013review,zheng2016person}, it is important to consider the neighbour images of the query image to obtain reliable multi-label. Therefore, GSMLP computes the similarity of the adjacent node distribution between $a_{i}$ and all others in $\hat{\mathcal{A}}$ and sorts the results as follows:
\begin{equation}
\begin{aligned}
&\text{Q}_{i} = \mathop{\arg \operatorname {sort}}_j dist(a_{i},a_{j}), \quad \text{w.r.t.,} 1 \leq j \leq n\\
&\text{Q}^{+}_{i} \leftarrow{} \text{Q}[1:|P^{+}_{i}|],
\label{eq:second} 
\end{aligned}
\end{equation}
where $\text{Q}_{i}$ is the label list sorted in descending order by the adjacent node similarity with respect to the image $x_{i}$. $dist(\cdot)$ indicates the Euclidean distance function. $|\text{P}^{+}_{i}|$ is the cardinality of set $P^{+}_{i}$, and $\text{Q}^{+}_{i}$ is a refined set as the top-$|\text{P}^{+}_{i}|$ elements in $\text{Q}_{i}$. That is, we choose $|\text{P}^{+}_{i}|$ nodes with the highest adjacent node similarity order.

The positive label is defined by the duplicated elements in $\text{P}^{+}_{i}$ and $\text{Q}^{+}_{i}$. Otherwise, it is classified as negative label. Intuitively, the duplicate elements mean that they are positioned in the near distance and have more mutual neighbours. Consequently, the predicted multi-label is represented as
\begin{equation}
\bar{y}_{i,j} =
\begin{cases}
1, & j\in \text{P}^{+}_{i} \wedge \text{Q}^{+}_{i} \\
0, & otherwise,
\end{cases}
\quad \text{w.r.t.,} 1 \leq j \leq n.
\label{eq:mlp} 
\end{equation}

\begin{figure}[t]
	\centering
	\includegraphics[width=\columnwidth]{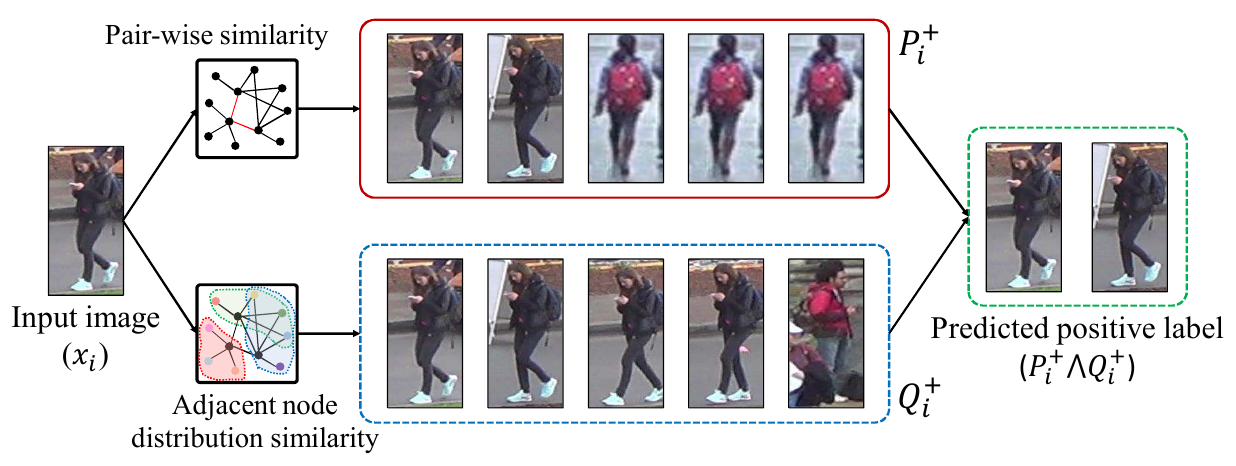}
	\caption{An illustration of multi-label prediction using GSMLP}
	\label{fig:comparison_labeling}
	\vspace{-2ex}
\end{figure}

In further section, the positive labels (i.e., $\bar{y}_{i,j}=1$) and the negative labels (i.e., $\bar{y}_{i,j}=0$) in the multi-label $\bar{y}_{i}$ are denoted by $\bar{y}^{+}_{i}$ and $\bar{y}^{-}_{i}$, respectively, for notation efficiency. Figure \ref{fig:comparison_labeling} shows how GSMLP predicts precise positive labels. We provide comprehensive ablation study about the multi-label prediction using GSMLP in Section \ref{sec:4:3}.

\subsection{Selective Multi-Label Classification loss}
\label{sec:3:3}
A commonly used approach to address multi-label classification is dividing a multi-label classification into the multiple single-class classifications \cite{Durand_2019_CVPR,lin2019bottom}. However, this approach has two challenging issues. First, since it needs to repeat single-class classification for all positive labels, it is computationally intensive. Second, usually, multi-labels are composed of many of negative labels and a few of positive labels, and this may suffer from a model collapse problem that a model is optimised into a wrong direction (\ie biased into negative labels \cite{wang2020unsupervised}). 

The proposed SMLC breaks through those challenging issues by reformulating a multi-label classification problem into a min-max problem for the distance between positive labels and negative labels, respectively. Since features, obtained from the embedding function $\phi$ or stored in the look-up table $\mathcal{M}$, are regularised by a unit vector, the values on $s_{i}$ are mapped into $[-1,1]$ (See Eq. \eqref{eq:sim_cosine}). Based on this value constraint, a loss term of SMLC is defined by
\begin{equation}
\begin{aligned}
\mathcal{L}(x_{i},\bar{y}_{i}) &= \sum_{j=0}^{n} \lVert{}s_{i,j}+(-1)^{\bar{y}_{i,j}}\lVert{}^{2},
\label{eq:loss_term} 
\end{aligned}
\end{equation}
where $s_{i,j}$ is 
the similarity between $z_{i}$ and $m_{j}$ computed by $\phi(x_{i}) \mathcal{M}^{\mathrm{T}}$. For positive labels ($\bar{y}^{+}_{i}$), our method optimises $s_{i,j}$ as 1, and for negative labels ($\bar{y}^{-}_{i}$), our method optimises $s_{i,j}$ as -1. In terms of cosine similarity, the meaning of the scheme is maximising similarities for positive labels and minimising similarities for negative labels. Compared with the aforementioned common approach, our approach is computationally less intensive because it does not need extra fully connected layer for the classification of every positive labels.

\textbf{Hard negative label mining:} However, the second issue is still problematic which is caused by the quantitative unbalance between  
positive and negative labels. To end this, SMLC presents a hard negative label mining to select more informative negative labels for training our method. Based on Eq. \eqref{eq:mlp}, our method basically defines the negative label list with the following condition:
\begin{equation}
\text{P}^{-}_{i} = \{j|j\notin \text{P}^{+}_{i} \wedge \text{Q}^{+}_{i} \quad \text{w.r.t.,} \ 1 \leq j \leq n\}. 
\label{eq:condition_for_neg} 
\end{equation}
The hard negative labels can be interpreted as images that look similar to a query image but have different identity. We hence sort the negative label list $\text{P}^{-}_{i}$ in descending order by the similarity $s$ and select top-$\gamma$\% of negative labels as the hard negative labels as follows: 
\begin{equation}
\begin{aligned}
&\dot{y}^{-}_{i} = \mathop{\arg \operatorname {sort}}_{j\in\text{P}^{-}_{i}} s_{i,j},\\
&\ddot{y}^{-}_{i} \leftarrow{} \dot{y}^{-}_{i}[1:\lceil\gamma|\dot{y}^{-}_{i}|\rceil],
\label{eq:mining_hard_label} 
\end{aligned}
\end{equation}
where $\dot{y}^{-}_{i}$ and $\ddot{y}^{-}_{i}$ are the sorted set of negative labels and the set of selected negative labels as hard negative labels, respectively. $|\dot{y}^{-}_{i}|$ is the cardinality of $\dot{y}^{-}_{i}$, and $\lceil\cdot\rceil$ denotes ceiling function.

Consequently, given the positive labels $\bar{y}^{+}_{i}$ and the selected hard negative labels $\ddot{y}^{-}_{i}$, the proposed method using SMLC optimises the follow:
\begin{equation}
\begin{aligned}
\mathcal{L}_{\text{SMLC}} &= \mathbb{E}_{\bar{y}^{+}_{i}\subset\bar{y}_{i}}[\mathcal{L}(x_{i},\bar{y}^{+}_{i})]+\mathbb{E}_{\ddot{y}^{-}_{i}\subset\bar{y}_{i}}[\mathcal{L}(x_{i},\ddot{y}^{-}_{i})].
\label{eq:min_max_representation} 
\end{aligned}
\end{equation}
$\gamma$ in SMLC can affect to the Re-ID performance by deciding the number of hard negative labels. We will test the effect of $\gamma$ in our experiments. 

\begin{figure*}
    \centering
    \begin{subfigure}{0.49\linewidth}
        \includegraphics[width=\linewidth]{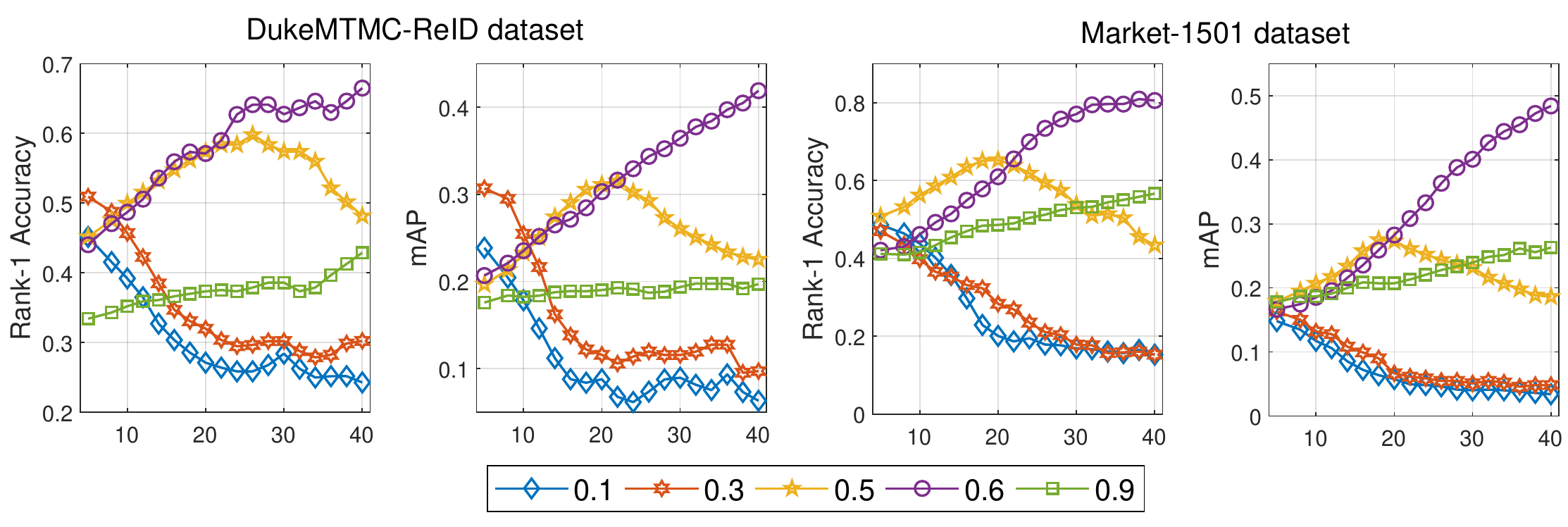}
    \caption{Experimental results on $\tau$}
    \end{subfigure}
    \begin{subfigure}{0.49\linewidth}
        \includegraphics[width=\linewidth]{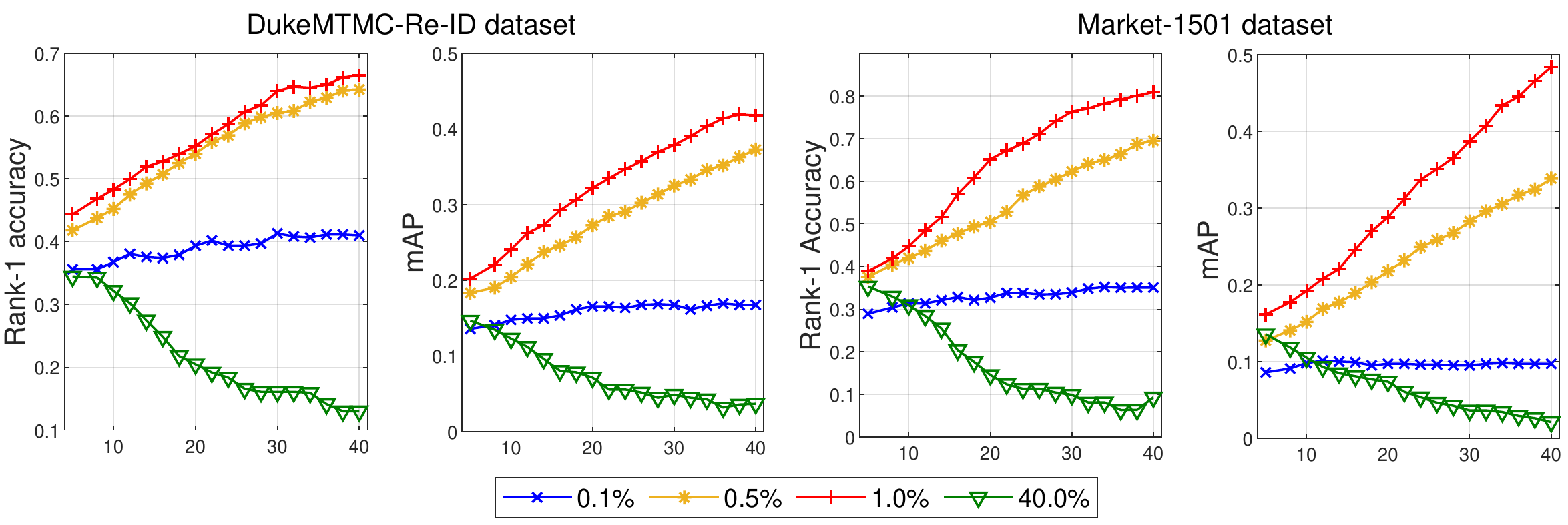}
    \caption{Experimental results on $\gamma$}
    \end{subfigure}
    \vspace{-1ex}
    \caption{Ablation studies for $\tau$ in GSMLP and $\gamma$ in SMLC. (a) and (b) represents the trends of rank-1 accuracies and mAPs at each training epoch, with respect to the value of $\tau$ and $\gamma$, on \textit{DukeMTMC-Re-ID} and \textit{Market-1501} datasets.}
\label{fig:as_tau_gamma}   
\vspace{-2ex} 
\end{figure*}

\section{Experiments}
\label{sec:4}
\subsection{Datasets and Evaluation Metrics}
\label{sec:4:1}
Three publicly available datasets, \textit{Market-1501} \cite{zheng2015scalable}, \textit{DukeMTMC-Re-ID} \cite{ristani2016performance}, and \textit{MSMT17} \cite{wei2018person}, are exploited in our experiments. Key properties of the three dataset are described in Table \ref{tbl:dataset_explain}. In particular, \textit{MSMT17} suffers from substantial variations of scene and illumination conditions, and those issues can be regarded as a challenging problem compared with the other two datasets. Our experiments for unsupervised person Re-ID have conducted with the standard protocols \cite{zheng2015scalable,ristani2016performance,wei2018person}. Cumulative Matching Characteristics (CMC) and Mean Average Precision (mAP) are used as evaluation metrics. 

\subsection{Implementation}
\label{sec:4:2}
ResNet-50 \cite{he2016deep} pretrained by ImageNet \cite{krizhevsky2012imagenet} is employed as the backbone of the embedding function $\phi$. The output of the $5^{\text{th}}$ pooling layer is used as features $z$, and the dimensionality of the output is 2,048. Simple data augmentation, such as random crop, rotation, and colour jitters, are used to improve the generalisation performance of learnt features. At the beginning of the training, the look-up table is initialised by 0, and GSMLP is conducted after 5 epochs to ensure the minimum qualities of the predicted multi-labels. Before using the predicted multi-labels, our method is trained using the single-class labels $y$. 

All images are resized to 256$\times$128. All models are optimized using Stochastic Gradient Descent (SGD) with a momentum of 0.9 for 40 epochs. The initial learning rates of the embedding function $\phi$ is 0.01. The learning rates are decayed by multiplying 0.1 for every 10 epoch, and the size of batch is 128. $\mathcal{M}$ is completely reinitialised every 5-epoch in the training step. The values of the hyper-parameters $\tau$ and $\gamma$ are fixed to 0.6 and 0.1, respectively. We provide insights to decide these two hyper-parameters in Section \ref{sec:4:3}. We implement our method using Pytorch, and all experiments are carried out with GTX2080Ti.

\begin{table}
\resizebox{\columnwidth}{!}{%
\begin{tabular}{l|c|c|c}
\hline
Dataset & Identities & Images & Cameras   \\
\hline\hline
Market-1501 \cite{zheng2015scalable} & 1,501 & 32,668 & 6    \\
DukeMTMC-Re-ID \cite{ristani2016performance} & 1,812 & 36,411 & 8  \\
MSMT17 \cite{wei2018person} & 4,101 & 126,441 & 15  \\
\hline
\end{tabular}
}
\caption{Key properties of the person Re-ID datasets. `Identities', `Images', and `Cameras' denotes the number of identities, images, and cameras of each dataset, respectively.}
\vspace{-2ex}
\label{tbl:dataset_explain}
\end{table}

\subsection{Ablation Study}
\label{sec:4:3}
We evaluate unsupervised person Re-ID performance depending on the setting of $\tau$ and $\gamma$. Additionally, we demonstrate the effectiveness of GSMLP and SMLC. All experiments are conducted with unsupervised Re-ID setting on \textit{Market-1501} and \textit{DukeMTMC-Re-ID}. Parameters that are not subject to testing are fixed during the experiments. 

\textbf{Parameter analysis on $\tau$}: $\tau$ in GSMLP decides the sparsity of the softened adjacent matrix $\hat{\mathcal{A}}$. The lower $\tau$ constructs the more dense graph. In other words, $\tau$ decides how many nodes need to be considered as adjacent nodes in computing the similarity of neighbour node distribution (Eq. \eqref{eq:second}). As shown in Figure \ref{fig:as_tau_gamma}(a), too low $\tau$ causes performance degradation. The results (using $\tau$ of 0.1 and 0.3) show that there is substantial performance degradation. The rank-1 accuracy and mAP using $\tau$ of 0.5 is gradually improves and then degrades again. Those trends can be interpreted that many false-positive labels are predicted, and these badly affect the optimisation process. However, too large value ($\tau$=0.9) is also not enough to obtain the best performance, and it means that a lot of true-positive labels are probably classified into negative labels. The best performance is achieved by $\tau$ of 0.6, and this value would be fixed in further experiments.

\textbf{Parameter analysis on $\gamma$}: Intuitively, $\gamma$ in SMLC decides the number of negative labels in computing $\mathcal{L}_{\text{SMLC}}$. Figure \ref{fig:as_tau_gamma}(b) reports the performance analysis on $\gamma$ of SMLC. We observe the performance with regard to the value of $\gamma$. We take 0.1\%, 0.5\%, 1.0\% and 40\% of negative labels as hard negative labels. The results in Figure \ref{fig:as_tau_gamma}(b) show that considering too many or too few hard negative labels is harmful in boosting the performance of unsupervised person Re-ID model. Particularly, considering too many negative labels can converge the proposed method into negative samples, and it can degrade the person Re-ID performance significantly. The best performance is achieved by $\gamma$ of 1\%, and this setting would be preserved in further experiments.

\begin{figure}[t]
	\centering
	\includegraphics[width=\columnwidth]{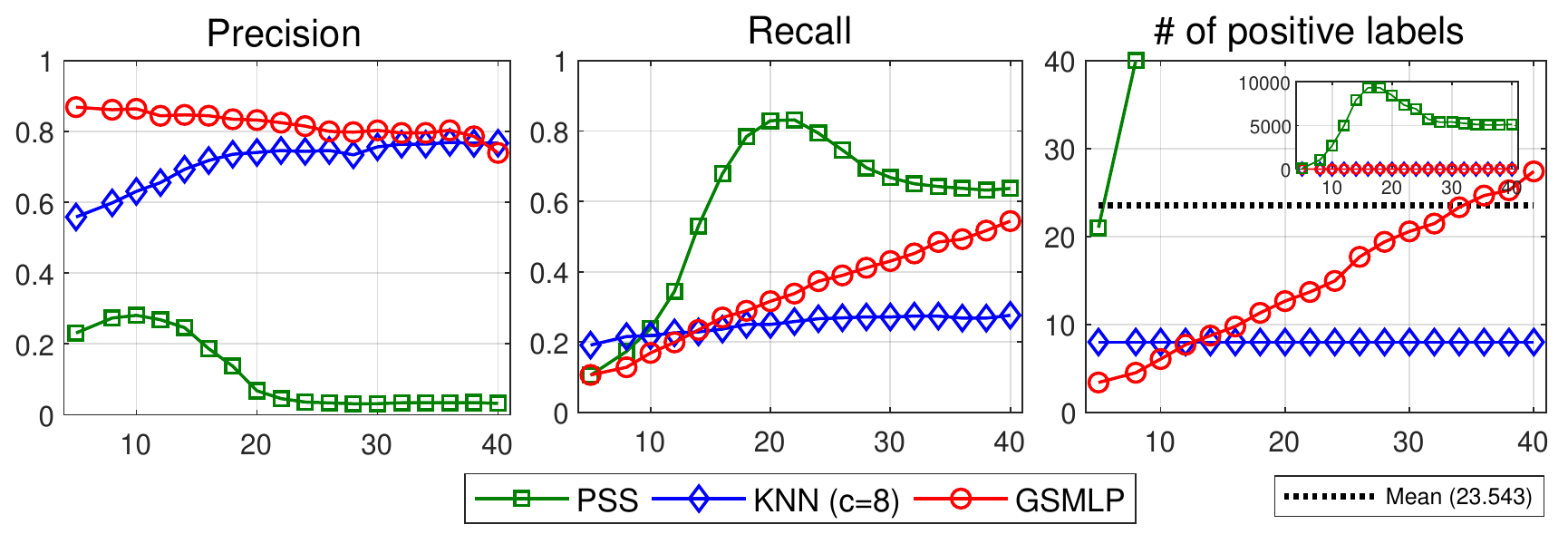}
	\caption{Evaluation of the multi-label prediction performances of PSS, KNN ($c$=8), and the proposed GSMLP. Precision, recall, and the amount of predicted positive labels according to the training epoch are shown. \textit{DukeMTMC-Re-ID} is used on this experiments.}
	\label{fig:mul_performance}
	\vspace{-2ex}
\end{figure}

\textbf{Performance analysis of GSMLP}: To demonstrate the effectiveness of GSMLP in predicting multi-label, we compare GSMLP with k-Nearest Neighbour (KNN) and pairwise-similarity score (PSS) based multi-label predictions (MLPs). Figure \ref{fig:mul_performance} shows that GSMLP finds positive labels increasingly. As the number of positive labels increases, the precision is gradually decreased, but in the overall training period, the precision of GSMLP is higher than other methods and the recall has been improved. PSS achieves the highest recall curve, but its precision is extremely lower than that of others since the number of positive labels is too many in which contains a lot of false-positive outcomes. KNN obtains comparable precision curve to the GSMLP, but the recall is the lowest among the three methods since the number of predictable labels is restricted. This means KNN may unable to provide enough quantity of positive labels. Additionally, we apply the predicted positive labels to train our method using SMLC. As shown in Table \ref{tbl:comparison-loss&mlp}. SMLC with GSMLP obtains the rank-1 accuracy of 80.9 and mAP of 48.4 on \textit{Market-1501} and the rank-1 accuracy of 66.2 and mAP of 41.4 on \textit{DukeMTMC-Re-ID}. These figures are higher than those obtained from other methods. 

\textbf{Performance analysis of SMLC}: SMLC is proposed to provide a stable learning process for multi-label classification which highly biased to negative labels. Based on above predicted multi-labels from the three methods, we train our method using SMLC and cross-entropy. In addition to the performance comparison on MLP approaches, Table \ref{tbl:comparison-loss&mlp} also shows the rank-1 accuracy and mAP depending on the objective function. The best performance is achieved by the model trained by SMLC with the multi-labels predicted by GSMLP. It produces the rank-1 accuracy of 80.09 and the mAP of 48.4 on \textit{Market-1501} dataset, and the rank-1 accuracy of 66.2 and the mAP of 41.4 on \textit{DukeMTMC-Re-ID} dataset. Those figures are higher than them achieved by the model trained by the cross-entropy using GSMLP. Overall results demonstrate that SMLC produces superior performance compared with the cross-entropy. Entire experimental results on Table \ref{tbl:comparison-loss&mlp} justify the effectiveness of SMLC for multi-label classification.

\begin{table}
\resizebox{\columnwidth}{!}{%
\begin{tabular}{l||l||c|c||c|c}
\hline
\multirow{2}{*}{Loss function} & \multirow{2}{*}{MLP approach} & \multicolumn{2}{c||}{Market-1501} & \multicolumn{2}{c}{DukeMTMC-Re-ID}  \\
\cline{3-4} \cline{5-6}
& & Rank-1 & mAP & Rank-1 & mAP  \\
\hline
\multirow{4}{*}{SMLC} & PSS & 51.1 & 23.3 & 55.4 & 25.6 \\
&KNN (c=4)  & 71.3 & 33.6 & 58.9 & 34.1 \\
&KNN (c=8)  & 72.9 & 35.1 & 60.1 & 35.7 \\
\cline{2-6}
&GSMLP & \textbf{80.9} & \textbf{48.4} & \textbf{66.2}& \textbf{41.4} \\
\hline\hline
\multirow{4}{*}{Cross-Entropy} & PSS & 47.8 & 21.7 & 50.2 & 24.2 \\
&KNN (c=4)  & 62.1 & 32.7 & 53.3 & 30.2 \\
&KNN (c=8)  & 64.7 & 36.1 & 55.8 & 31.8\\
\cline{2-6}
&GSMLP & 71.4 & 51.0 & 60.0 & 37.8 \\
\hline
\end{tabular}
}
\caption{Quantitative performance comparison of different loss functions and multi-label prediction (MLP) approaches. 4 and 8 represent the number of centroids for KNN algorithms.} 
\vspace{-2ex}
\label{tbl:comparison-loss&mlp}
\end{table}

\begin{table*}
\resizebox{\textwidth}{!}{%
\setlength{\tabcolsep}{2.5mm}{
\begin{tabular}{l||c||c|c|c|c|c||c|c|c|c|c}
\hline
\multirow{2}{*}{Method} & \multirow{2}{*}{Approach} & \multicolumn{5}{c||}{Market-1501} & \multicolumn{5}{c}{DukeMTMC-Re-ID} \\
\cline{3-12}
&  & Source & Rank-1 & Rank-5 & Rank-10 & mAP & Source & Rank-1 & Rank-5 & Rank-10 & mAP \\
\hline\hline
PUL\cite{fan2018unsupervised} & Domain adapted & Duke & 45.5 &60.7 &66.7& 20.5& Market & 30.0 &43.4 &48.5 &16.4 \\
PTGAN\cite{wei2018person} &Domain adapted& Duke & 38.6& -& 66.1& -& Market & 27.4& - &50.7 & - \\
TJ-AIDL\cite{wang2018transferable} & Domain adapted & Duke &58.2 &74.8 &81.1 &26.5& Market & 44.3& 59.6 &65.0& 23.0 \\
ECN\cite{zhong2019invariance} & Domain adapted & Duke & 75.1 & 87.6 & 91.6 & 43.0 &Market & 63.3 & 75.8 & 80.4 & 40.4 \\
MAR\cite{yu2019unsupervised}  &Domain adapted & MSMT & 67.7  &81.9  &-  &40.0 & MSMT & 67.1 & 79.8  & -  &48.0\\
PAUL\cite{Yang_2019_CVPR} & Domain adapted & MSMT & 68.5 & 82.4 & 87.4 & 40.1 & MSMT & 72.0 & 82.7 & 86.0 & 53.2 \\
SSG\cite{Fu_2019_ICCV} & Domain adapted & Duke & 80.0 & 90.0 & 92.4 & 58.3 & Market & 73.0 & 80.6 & 83.2 & 53.4 \\
CASCL\cite{Wu_2019_ICCV} & Domain adapted & MSMT & 65.4 & 80.6 & 86.2 & 35.5 & MSMT & 59.3 & 73.2 & 77.5 & 37.8\\
PAST\cite{Zhang_2019_ICCV} & Domain adapted & Duke & 78.4 & - & - & 54.6 & Market & 72.4 & - & - & 54.3\\
GDS-H \cite{jin2020global} & Domain adapted & Duke & 81.1 & - & - & 61.2 & Market & 73.1 & - & - & 55.1\\
ADTC \cite{wu2020attention} & Domain adapted & Duke & 79.3 & 90.8 & 94.1 & 59.7 & Market & 71.9 & 84.1 & 87.5 & 52.5\\
DG-Net++ \cite{zou2020joint} & Domain adapted & Duke & 82.1 & 90.2 & 92.7 & 61.7 & Market & \underline{78.9} & \underline{87.8} & \underline{90.4} & \underline{63.8}\\
NRMT \cite{zhao2020unsupervised} & Domain adapted & Duke & \underline{87.8} & \underline{94.6} & \underline{96.5} & \underline{71.7} & Market & 77.8 & 86.9 & 89.5 & 62.2\\
AD-Cluster \cite{zhao2020unsupervised} & Domain adapted & Duke & 86.7 & 94.4 & \underline{96.5} & 68.3 & Market & 72.6 & 82.5 & 85.5 & 54.1\\
\hline
\hline
LOMO\cite{liao2015person} & Unsupervised & None &  27.2& 41.6& 49.1 &8.0& None & 12.3 &21.3& 26.6& 4.8 \\
BOW\cite{zheng2015scalable} & Unsupervised & None & 35.8& 52.4& 60.3 &14.8 & None &17.1& 28.8& 34.9 &8.3 \\
OIM\cite{xiao2017joint} & Unsupervised & None & 38.0 &  58.0 & 66.3 & 14.0 & None & 24.5 & 38.8 & 46.0 & 11.3 \\
BUC\cite{lin2019bottom} & Unsupervised & None & 66.2 & 79.6 & 84.5 & 38.3 & None & 47.4 & 62.6 & 68.4 & 27.5 \\
DBC\cite{ding12dispersion} & Unsupervised & None & 69.2 & 83.0 & 87.8 & 41.3 & None & 51.5 & 64.6 & 70.1 & 30.0 \\
SSL \cite{lin2020unsupervised} & Unsupervised & None & 71.7 & 83.8 & 87.4 & 37.8 & None & 52.5 & 63.5 & 68.9 & 28.6\\
MMCL \cite{wang2020unsupervised} & Unsupervised & None & 80.3 & 89.4 & 92.3 & 45.5 & None & 65.2 & 75.9 & 80.0 & 40.2\\
\hline
Our method & Unsupervised & None & \textbf{80.9} & \textbf{90.2} & \textbf{93.1} & \textbf{48.4} & None & \textbf{66.5} & \textbf{77.4} & \textbf{81.0} & \textbf{41.9} \\
\hline
\end{tabular}
}
}
\caption{Quantitative performance comparison for unsupervised person Re-ID with existing state-of-the-art methods including DA (\ie `Domain adapted') based methods. `Unsupervised' means the methods do not need pre-labelled data in the training step. The underlined figures represent the best performance among the DA based methods. The bolded figures indicate the best performance among the unsupervised methods that do not need labelled source dataset.}
\vspace{-2ex}
\label{tbl:comparison-market-duke}
\end{table*}

\begin{table}
\resizebox{\columnwidth}{!}{%
\begin{tabular}{l|c|c|c|c|c}
\hline
\multirow{2}{*}{Method} & \multirow{2}{*}{Setting} & \multicolumn{4}{c}{MSMT17} \\
\cline{3-6}
& &Rank-1 & Rank-5 & Rank-10 & mAP  \\
\hline\hline
PTGAN~\cite{wei2018person} & DA (Duke) &11.8 & - & 27.4 & 3.3 \\
ECN~\cite{zhong2019invariance} & DA (Duke) & 30.2 & 41.5 & 46.8 & 10.2 \\
SSG~\cite{Fu_2019_ICCV} & DA (Duke) & 32.2 & - & 51.2 & 13.3\\
DG-Net++~\cite{zou2020joint} & DA (Duke) & \underline{49.8} & \underline{60.9} & \underline{65.9} & \underline{22.1}\\
NRMT~\cite{zhao2020unsupervised} & DA (Duke) & 45.2 & 57.8 & 63.3 & 20.6\\
\hline\hline
MMCL~\cite{wang2020unsupervised} & Unsuper. & 35.4 & \textbf{44.8} & \textbf{49.8} & 11.2 \\
CCSE~\cite{lin2020unsupervised_2} & Unsuper. & 31.4 & 41.4 & 45.5 & 9.9 \\
\hline
Our method & Unsuper. & \textbf{36.2} & 44.0 & 48.1 & \textbf{15.8} \\
\hline
\end{tabular}
}
\caption{Comparison of unsupervised person Re-ID performance on \textit{MSMT17}. `DA (Duke)' denotes that their methods employ `domain adaptation' using \textit{DukeMTMC-Re-ID}. `Unsuper.' means that the method does not need any types of prepared labels.} 
\vspace{-2ex}
\label{tbl:comparison-msmt17}
\end{table}

\subsection{Comparison with state-of-the-art methods}
\label{sec:4:4}
Our method is compared with various recent state-of-the-art unsupervised Re-ID methods including DA based methods. Table \ref{tbl:comparison-market-duke} shows quantitative comparison on \textit{Market-1501} and \textit{DukeMTMC-Re-ID}. On the comparison with the unsupervised methods, the methods, LOMO \cite{liao2015person} and BOW \cite{zheng2015scalable}, based on hand-crafted features produce lower performance than others. BUC \cite{lin2019bottom} and DBC \cite{ding12dispersion} define each image as a single cluster. Our method outperforms BUC and DBC with large margins. This performance gap can be interpreted as follows. First, their performance can be variant to the positions of initial clusters. This issue is one of the inevitable issues for all clustering-based methods. Second, those methods do not consider the number of data assigned to each cluster, which means unbalance between each positive label can be caused.

Our method also achieves better performance than SSL \cite{lin2020unsupervised} and MMCL \cite{wang2020unsupervised} which are the latest proposed methods for unsupervised person Re-ID. Both methods initially assign the file indices as a single-class label and create multi-labels to represent the relevance of other images that may belong to the same identity. Accordingly, those methods are methodologically similar to our work. The difference in performance can be interpreted as follows. In generating the multi-labels, SSL only compares a pair-wise similarity of features, so it may create lots of false-positive results. MMCL takes into account a cycle consistency, but GSMLP on our method can check cycle consistency and the similarities of each features' neighbour simultaneously. 

In the comparison with the methods based on DA, our method produces comparable performances to the state-of-the-art methods. For example, Rank-1 accuracy of our method on \textit{Market-1501} and \textit{DukeMTMC-Re-ID} are 80.9 and 66.5, respectively. Our method achieves the mAP of 48.4 and 41.9 for the two datasets. On the experiments using \textit{Market-1501}, this figure is clearly higher than that of several approaches \cite{fan2018unsupervised,wei2018person,wang2018transferable,zhong2019invariance} and partially better than that of SSG\cite{Fu_2019_ICCV}, PAUL\cite{Yang_2019_CVPR}, ADTC \cite{wu2020attention}, and PAST\cite{Zhang_2019_ICCV}. The experimental results can be interpreted that using a labelled dataset is a clear advantage in deriving discriminative person Re-ID model. Nevertheless, although several approaches achieve higher performance than our method, our method can be considered as a more flexible solution because our method does not need any pre-labelled dataset. 

Additionally, a comparison on \textit{MSMT17} (see Table \ref{tbl:comparison-msmt17}) also shows that our method outperforms CCSE \cite{lin2020unsupervised_2} and is also compatible to the methods using DA \cite{wei2018person,zhong2019invariance,Fu_2019_ICCV,zou2020joint,zhao2020unsupervised}. Our method achieves 36.2 of Rank-1 accuracy and 15.8 of mAP which is the best performance among the unsupervised methods. Those figures are also comparable to the methods based on DA. Consequently, overall results on the three datasets demonstrate GSMLP and SMLC can provide promising performances under the unsupervised manner for person Re-ID.

\section{Conclusion}
\label{sec:5}
This paper has proposed a multi-label prediction and classification based on graph structural insight to handle unsupervised person Re-ID. The proposed GSMLP has predicted good quality multi-labels by analysing features and their neighbours based on graph structural insight. In addition, the proposed SMLC has provided stable learning process about multi-label classification with less computational costs, and it has reduced risk about imbalance problem between positive and negative labels. Experimental results has demonstrated the efficiencies of GSMLP and SMLC. In comparison with the existing state-of-the-art methods, our method has outperformed other unsupervised methods and shown the competitive performance compared with the methods based on DA that need labelled source datasets.

{\small
\bibliographystyle{ieee_fullname}
\bibliography{egpaper_for_review.bbl}

\begin{thebibliography}{10}\itemsep=-1pt

\bibitem{40d5d7fd62cb44ba934a8a75d4b2b076}
Yoshua Bengio and Yann Lecun.
\newblock {\em Scaling learning algorithms towards AI}.
\newblock MIT Press, 2007.

\bibitem{chen2020salience}
Xuesong Chen, Canmiao Fu, Yong Zhao, Feng Zheng, Jingkuan Song, Rongrong Ji,
  and Yi Yang.
\newblock Salience-guided cascaded suppression network for person
  re-identification.
\newblock In {\em Proceedings of the IEEE/CVF Conference on Computer Vision and
  Pattern Recognition}, pages 3300--3310, 2020.

\bibitem{ding12dispersion}
Guodong Ding, Salman Khan, Qingze Yin, and Zhenmin Tang.
\newblock Dispersion based clustering for unsupervised person
  re-identification.
\newblock In {\em BMVC}, 2019.

\bibitem{Durand_2019_CVPR}
Thibaut Durand, Nazanin Mehrasa, and Greg Mori.
\newblock Learning a deep convnet for multi-label classification with partial
  labels.
\newblock In {\em CVPR}, 2019.

\bibitem{fan2018unsupervised}
Hehe Fan, Liang Zheng, Chenggang Yan, and Yi Yang.
\newblock Unsupervised person re-identification: Clustering and fine-tuning.
\newblock {\em ACM Transactions on Multimedia Computing, Communications, and
  Applications (TOMM)}, 14(4):83, 2018.

\bibitem{Fu_2019_ICCV}
Yang Fu, Yunchao Wei, Guanshuo Wang, Yuqian Zhou, Honghui Shi, and Thomas~S.
  Huang.
\newblock Self-similarity grouping: A simple unsupervised cross domain
  adaptation approach for person re-identification.
\newblock In {\em ICCV}, 2019.

\bibitem{gray2008viewpoint}
Douglas Gray and Hai Tao.
\newblock Viewpoint invariant pedestrian recognition with an ensemble of
  localized features.
\newblock In {\em European conference on computer vision}, pages 262--275.
  Springer, 2008.

\bibitem{he2016deep}
Kaiming He, Xiangyu Zhang, Shaoqing Ren, and Jian Sun.
\newblock Deep residual learning for image recognition.
\newblock In {\em CVPR}, 2016.

\bibitem{he2020guided}
Lingxiao He and Wu Liu.
\newblock Guided saliency feature learning for person re-identification in
  crowded scenes.
\newblock In {\em European Conference on Computer Vision}, pages 357--373.
  Springer, 2020.

\bibitem{jegou2007contextual}
Herve Jegou, Hedi Harzallah, and Cordelia Schmid.
\newblock A contextual dissimilarity measure for accurate and efficient image
  search.
\newblock In {\em CVPR}, 2007.

\bibitem{jin2020global}
Xin Jin, Cuiling Lan, Wenjun Zeng, and Zhibo Chen.
\newblock Global distance-distributions separation for unsupervised person
  re-identification.
\newblock 2020.

\bibitem{BMVC2015_44}
Elyor Kodirov, Tao Xiang, and Shaogang Gong.
\newblock Dictionary learning with iterative laplacian regularisation for
  unsupervised person re-identification.
\newblock In {\em Proceedings of the British Machine Vision Conference (BMVC)},
  pages 44.1--44.12. BMVA Press, September 2015.

\bibitem{krizhevsky2012imagenet}
Alex Krizhevsky, Ilya Sutskever, and Geoffrey~E Hinton.
\newblock Imagenet classification with deep convolutional neural networks.
\newblock In {\em NeurIPS}, 2012.

\bibitem{li2017person}
Sheng Li, Ming Shao, and Yun Fu.
\newblock Person re-identification by cross-view multi-level dictionary
  learning.
\newblock {\em IEEE transactions on pattern analysis and machine intelligence},
  40(12):2963--2977, 2017.

\bibitem{li2014deepreid}
Wei Li, Rui Zhao, Tong Xiao, and Xiaogang Wang.
\newblock Deepreid: Deep filter pairing neural network for person
  re-identification.
\newblock In {\em Proceedings of the IEEE conference on computer vision and
  pattern recognition}, pages 152--159, 2014.

\bibitem{liao2015person}
Shengcai Liao, Yang Hu, Xiangyu Zhu, and Stan~Z Li.
\newblock Person re-identification by local maximal occurrence representation
  and metric learning.
\newblock In {\em CVPR}, 2015.

\bibitem{lin2019bottom}
Yutian Lin, Xuanyi Dong, Liang Zheng, Yan Yan, and Yi Yang.
\newblock A bottom-up clustering approach to unsupervised person
  re-identification.
\newblock In {\em AAAI}, 2019.

\bibitem{lin2020unsupervised_2}
Yutian Lin, Yu Wu, Chenggang Yan, Mingliang Xu, and Yi Yang.
\newblock Unsupervised person re-identification via cross-camera similarity
  exploration.
\newblock {\em IEEE Transactions on Image Processing}, 29:5481--5490, 2020.

\bibitem{lin2020unsupervised}
Yutian Lin, Lingxi Xie, Yu Wu, Chenggang Yan, and Qi Tian.
\newblock Unsupervised person re-identification via softened similarity
  learning.
\newblock In {\em Proceedings of the IEEE/CVF Conference on Computer Vision and
  Pattern Recognition}, pages 3390--3399, 2020.

\bibitem{long2015learning}
Mingsheng Long, Yue Cao, Jianmin Wang, and Michael~I Jordan.
\newblock Learning transferable features with deep adaptation networks.
\newblock {\em arXiv preprint arXiv:1502.02791}, 2015.

\bibitem{ristani2016performance}
Ergys Ristani, Francesco Solera, Roger Zou, Rita Cucchiara, and Carlo Tomasi.
\newblock Performance measures and a data set for multi-target, multi-camera
  tracking.
\newblock In {\em ECCV}, 2016.

\bibitem{schneider2020improving}
Steffen Schneider, Evgenia Rusak, Luisa Eck, Oliver Bringmann, Wieland Brendel,
  and Matthias Bethge.
\newblock Improving robustness against common corruptions by covariate shift
  adaptation.
\newblock {\em Advances in Neural Information Processing Systems}, 33, 2020.

\bibitem{su2017pose}
Chi Su, Jianing Li, Shiliang Zhang, Junliang Xing, Wen Gao, and Qi Tian.
\newblock Pose-driven deep convolutional model for person re-identification.
\newblock In {\em ICCV}, 2017.

\bibitem{varior2016gated}
Rahul~Rama Varior, Mrinal Haloi, and Gang Wang.
\newblock Gated siamese convolutional neural network architecture for human
  re-identification.
\newblock In {\em European conference on computer vision}, pages 791--808.
  Springer, 2016.

\bibitem{wang2020unsupervised}
Dongkai Wang and Shiliang Zhang.
\newblock Unsupervised person re-identification via multi-label classification.
\newblock In {\em Proceedings of the IEEE/CVF Conference on Computer Vision and
  Pattern Recognition}, pages 10981--10990, 2020.

\bibitem{wang2018transferable}
Jingya Wang, Xiatian Zhu, Shaogang Gong, and Wei Li.
\newblock Transferable joint attribute-identity deep learning for unsupervised
  person re-identification.
\newblock In {\em CVPR}, 2018.

\bibitem{wei2018person}
Longhui Wei, Shiliang Zhang, Wen Gao, and Qi Tian.
\newblock Person transfer gan to bridge domain gap for person
  re-identification.
\newblock In {\em CVPR}, 2018.

\bibitem{Wu_2019_ICCV}
Ancong Wu, Wei-Shi Zheng, and Jian-Huang Lai.
\newblock Unsupervised person re-identification by camera-aware similarity
  consistency learning.
\newblock In {\em ICCV}, 2019.

\bibitem{wu2020attention}
Si Wu.
\newblock An attention-driven two-stage clustering method for unsupervised
  person re-identification.
\newblock 2020.

\bibitem{xiao2017joint}
Tong Xiao, Shuang Li, Bochao Wang, Liang Lin, and Xiaogang Wang.
\newblock Joint detection and identification feature learning for person
  search.
\newblock In {\em Proceedings of the IEEE Conference on Computer Vision and
  Pattern Recognition}, pages 3415--3424, 2017.

\bibitem{Yan_2017_CVPR}
Hongliang Yan, Yukang Ding, Peihua Li, Qilong Wang, Yong Xu, and Wangmeng Zuo.
\newblock Mind the class weight bias: Weighted maximum mean discrepancy for
  unsupervised domain adaptation.
\newblock In {\em CVPR}, 2017.

\bibitem{Yang_2019_CVPR}
Qize Yang, Hong-Xing Yu, Ancong Wu, and Wei-Shi Zheng.
\newblock Patch-based discriminative feature learning for unsupervised person
  re-identification.
\newblock In {\em CVPR}, 2019.

\bibitem{yu2019unsupervised}
Hong-Xing Yu, Wei-Shi Zheng, Ancong Wu, Xiaowei Guo, Shaogang Gong, and
  Jian-Huang Lai.
\newblock Unsupervised person re-identification by soft multilabel learning.
\newblock In {\em CVPR}, 2019.

\bibitem{yu2018deep}
Jongmin Yu, Donghwuy Ko, Hangyul Moon, and Moongu Jeon.
\newblock Deep discriminative representation learning for face verification and
  person re-identification on unconstrained condition.
\newblock In {\em 2018 25th IEEE International Conference on Image Processing
  (ICIP)}, pages 1658--1662. IEEE, 2018.

\bibitem{zhang2013review}
Min-Ling Zhang and Zhi-Hua Zhou.
\newblock A review on multi-label learning algorithms.
\newblock {\em IEEE transactions on knowledge and data engineering},
  26(8):1819--1837, 2013.

\bibitem{Zhang_2019_ICCV}
Xinyu Zhang, Jiewei Cao, Chunhua Shen, and Mingyu You.
\newblock Self-training with progressive augmentation for unsupervised
  cross-domain person re-identification.
\newblock In {\em ICCV}, 2019.

\bibitem{zhao2020unsupervised}
Fang Zhao, Shengcai Liao, Guo-Sen Xie, Jian Zhao, Kaihao Zhang, and Ling Shao.
\newblock Unsupervised domain adaptation with noise resistible mutual-training
  for person re-identification.
\newblock In {\em European Conference on Computer Vision}, pages 526--544.
  Springer, 2020.

\bibitem{zheng2015scalable}
Liang Zheng, Liyue Shen, Lu Tian, Shengjin Wang, Jingdong Wang, and Qi Tian.
\newblock Scalable person re-identification: A benchmark.
\newblock In {\em Proceedings of the IEEE international conference on computer
  vision}, pages 1116--1124, 2015.

\bibitem{zheng2016person}
Liang Zheng, Yi Yang, and Alexander~G Hauptmann.
\newblock Person re-identification: Past, present and future.
\newblock {\em arXiv preprint arXiv:1610.02984}, 2016.

\bibitem{zheng2019joint}
Zhedong Zheng, Xiaodong Yang, Zhiding Yu, Liang Zheng, Yi Yang, and Jan Kautz.
\newblock Joint discriminative and generative learning for person
  re-identification.
\newblock In {\em Proceedings of the IEEE conference on computer vision and
  pattern recognition}, pages 2138--2147, 2019.

\bibitem{zhong2019invariance}
Zhun Zhong, Liang Zheng, Zhiming Luo, Shaozi Li, and Yi Yang.
\newblock Invariance matters: Exemplar memory for domain adaptive person
  re-identification.
\newblock In {\em CVPR}, 2019.

\bibitem{zhou2020online}
Jiahuan Zhou, Bing Su, and Ying Wu.
\newblock Online joint multi-metric adaptation from frequent sharing-subset
  mining for person re-identification.
\newblock In {\em Proceedings of the IEEE/CVF Conference on Computer Vision and
  Pattern Recognition}, pages 2909--2918, 2020.

\bibitem{zhou2020fine}
Qinqin Zhou, Bineng Zhong, Xiangyuan Lan, Gan Sun, Yulun Zhang, Baochang Zhang,
  and Rongrong Ji.
\newblock Fine-grained spatial alignment model for person re-identification
  with focal triplet loss.
\newblock {\em IEEE Transactions on Image Processing}, 29:7578--7589, 2020.

\bibitem{zou2020joint}
Yang Zou, Xiaodong Yang, Zhiding Yu, B.~V. K.~Vijaya Kumar, and Jan Kautz.
\newblock Joint disentangling and adaptation for cross-domain person
  re-identification, 2020.

\end{thebibliography}
}

\end{document}